\documentclass[letterpaper, 10 pt, journal, twoside]{IEEEtran}
\usepackage{amsmath,amsfonts}
\usepackage{algorithmic}
\usepackage{algorithm}
\usepackage{array}
\usepackage[caption=false,font=normalsize,labelfont=sf,textfont=sf]{subfig}
\usepackage{textcomp}
\usepackage{stfloats}
\usepackage{url}
\usepackage{verbatim}
\usepackage{graphicx}
\usepackage{xcolor}
\usepackage{cite}
\usepackage{booktabs}
\usepackage{multirow}
\usepackage[table]{xcolor}
\usepackage{colortbl}
\usepackage{array}
\usepackage{etoolbox}
\def\tablename{\textbf{Table}}
\makeatletter
\patchcmd{\@makecaption}{\scshape}{}{}{}
\makeatletter
\patchcmd{\@makecaption}{\\}{.\ }{}{}
\makeatother
\usepackage{threeparttable}
\usepackage[numbers,sort&compress]{natbib}
\usepackage{xcolor}
\usepackage{dashrule}
\usepackage{tikz}

\definecolor{Gray}{gray}{0.9}
\definecolor{LightOrange}{RGB}{255, 242, 224}
\begin{document}

\title{TakeAD: Preference-based Post-optimization for End-to-end Autonomous Driving with Expert Takeover Data}

\author{
${\text{Deqing Liu}}^{1*}$, $\text{Yinfeng Gao}^{2*}$, $\text{Deheng Qian}^{3}$, $\text{Qichao Zhang}^{1\ddag}$, $\text{Xiaoqing Ye}^{3}$, $\text{Junyu Han}^{3}$, $\text{Yupeng Zheng}^{1}$, $\text{Xueyi Liu}^{1}$, $\text{Zhongpu Xia}^{1}$, $\text{Dawei Ding}^{2}$, $\text{Yifeng Pan}^{3}$, and $\text{Dongbin Zhao}^{1}$, $~\IEEEmembership{Fellow, IEEE}$

\thanks{Manuscript received: July, 8, 2025; Revised September, 15, 2025; Accepted November, 24, 2025.}
\thanks{This paper was recommended for publication by
Editor Aleksandra Faust upon evaluation of the Associate Editor and Reviewers’ comments. 
This work is supported by National Key Research and Development Program of China under Grant 2022YFA1004000,  in part by Beijing Nova Program (20240484562) and Beijing Natural Science Foundation under Grant 4242052.}
\thanks{$^{1}$ Deqing Liu, Qichao Zhang, Yupeng Zheng, Xueyi Liu, Zhongpu Xia and Dongbin Zhao are with The State Key Laboratory of Multimodal Artificial Intelligence Systems, Institute of Automation, Chinese Academy of Sciences, Beijing 100190, China, and also with the School of Artificial Intelligence, University of Chinese Academy of Sciences, Beijing 100049, China. {\tt\footnotesize liudeqing2025@ia.ac.cn}}
\thanks{$^{2}$ Yinfeng Gao and Dawei Ding are with the School of Automation and Electrical Engineering, University of Science and Technology Beijing, Beijing 100083, China. {\tt\footnotesize gaoyinfeng07@gmail.com}
}
\thanks{$^{3}$ Deheng Qian, Xiaoqing Ye, Junyu Han and Yifeng Pan are with Chongqing Chang'an Technology Co., Ltd.}
\thanks{$^{*}$ Deqing Liu and Yinfeng Gao contribute equally to this work. $^{\ddag}$~Qichao Zhang is the corresponding author.
{\tt\footnotesize zhangqichao2014@ia.ac.cn}}
\thanks{Digital Object Identifier (DOI): see top of this page.}
\vspace{-8mm}
}
\markboth{IEEE ROBOTICS AND AUTOMATION LETTERS. PREPRINT VERSION. ACCEPTED NOVEMBER 2025}%
{Liu \MakeLowercase{\textit{et al.}}: Post-optimization for E2E AD with Takeover}


\maketitle
\begin{abstract}
 Existing end-to-end autonomous driving methods typically rely on imitation learning (IL) but face a key challenge: the misalignment between open-loop training and closed-loop deployment. {
This misalignment often triggers driver-initiated takeovers and system disengagements during closed-loop execution.  How to leverage those expert takeover data from disengagement scenarios and effectively expand the IL policy’s capability presents a valuable yet unexplored challenge.  In this paper, we propose TakeAD, a novel preference-based post-optimization framework that fine-tunes the pre-trained IL policy with this disengagement data to enhance the closed-loop driving performance.} First, we design an efficient expert takeover data collection pipeline inspired by human takeover mechanisms in real-world autonomous driving systems.
 Then, this post-optimization framework integrates iterative Dataset Aggregation (DAgger) for imitation learning with Direct Preference Optimization (DPO) for preference alignment.
The DAgger stage {equips} the policy with fundamental capabilities to handle {disengagement} states through direct imitation of expert interventions. 
Subsequently, the DPO stage refines the policy’s behavior to better align with expert preferences in disengagement scenarios.
{Through multiple iterations, the policy progressively learns recovery strategies for {disengagement} states, thereby mitigating the open-loop gap.}
Experiments on the closed-loop Bench2Drive benchmark demonstrate our method's effectiveness compared with pure IL methods, with comprehensive ablations confirming the contribution of each component. 
\end{abstract}


\IEEEpeerreviewmaketitle
\section{Introduction}
\IEEEPARstart{E}{nd-to-end} autonomous driving (E2E AD) has recently emerged as a prominent research focus~\cite{review,uniad}. 
This approach directly maps sensor inputs to planning trajectories or control signals using a unified neural network, typically trained through imitation learning (IL) on large-scale human demonstrations~\cite{vad,uncad}.
While IL achieves strong open-loop performance, its closed-loop deployment suffers from compounding errors that push the policy into {disengagement} states, significantly degrading closed-loop planning performance~\cite{DAgger, reasonplan}. 
This fundamental discrepancy between open-loop training and closed-loop execution remains a critical challenge. {Recently, further fine-tuning of IL policies, referred to as post-optimization, has emerged as a promising direction to overcome the limitations of pure IL-based training~\cite{rad}.}
\begin{figure}[!t]
\centering
\includegraphics[width=0.38\textwidth]{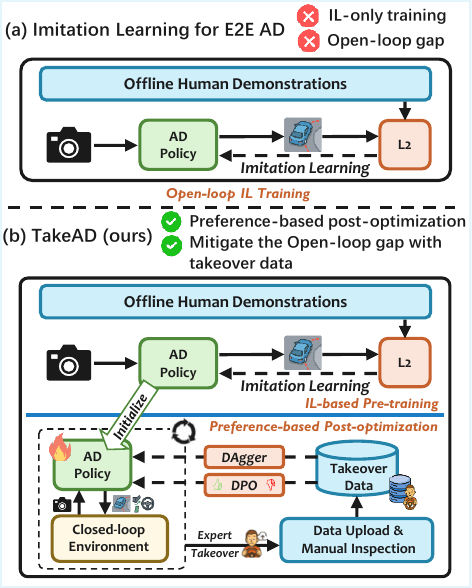}
\vspace{-3mm}
\caption{
{Different training paradigms of end-to-end autonomous driving. 
(a) Imitation learning matches human demonstrations but suffers from the open-loop gap and out-of-distribution problems. 
(b) The proposed TakeAD method, a novel preference-based post-optimization framework, leverages high-quality expert takeover data for iterative imitation and preference optimization.}}
\label{fig1}
\vspace{-6mm}
\end{figure}

During real-world deployment, when encountering edge cases that lead to safety-critical situations, autonomous vehicles require human driver takeovers. These disengagement cases are systematically recorded, 
offering valuable samples essential for further enhancing the driving policy. 
This raises a question: \textbf{\textit{How can these high-value takeover data be utilized to enhance the driving policy effectively?}}
Unfortunately, due to the lack of open-source takeover data, this issue has hardly been explored yet in the academic community.

To address these challenges, we propose \textbf{TakeAD}, a preference-based post-optimization framework for E2E AD that leverages expert {takeover} data.
First, we develop an automated pipeline for harvesting expert takeover data based on the current popular closed-loop Bench2Drive benchmark~\cite{b2d}. 
Specifically, during the closed-loop execution of an E2E AD policy, a {privileged} rule-based expert policy {(with access to perfect perception)} operates in parallel in shadow mode, intervening when abnormal actions or collision risks are detected, thereby generating high-quality takeover data from disengagement scenarios for post-training.
Second, we propose a preference-based post-optimization framework combining Dataset Aggregation (DAgger)~\cite{DAgger} and Direct Preference Optimization (DPO)~\cite{dpo} that effectively utilizes the collected takeover data, as shown in Fig. \ref{fig1}.
{As a classical online imitation learning algorithm, DAgger enables the policy to acquire fundamental capabilities for handling {disengagement} scenarios by directly imitating expert interventions. DPO, an efficient reinforcement learning from human feedback method without handcrafted rewards, further refines the policy’s behavior to align with expert preferences in disengagement situations. Iterative training progressively pushes the limits of the policy's capabilities in handling {challenging} driving scenarios.}
{Third, we introduce a reactive control branch that directly predicts low-level control actions (throttle, brake, and steering) at each timestep, allowing direct and immediate adjustments to the ego vehicle. Combining trajectory and control outputs enables the policy to better handle both general and interactive scenarios, leading to more robust closed-loop planning performance.}

Our main contributions are summarized as follows:
\begin{enumerate}
\item{{We propose a preference-based post-optimization framework TakeAD, which uses high-quality expert takeover data and integrates iterative DAgger and DPO. DAgger enables the driving policy with the initial capability to handle {disengagement} states, while DPO refines the policy's behavior to better align with expert preferences.}}
\item{{We introduce an efficient expert takeover data collection pipeline tailored for disengagement scenarios. 
During closed-loop policy execution, an expert policy runs in shadow mode and triggers a takeover when certain conditions are met, logging data for post-optimization.}}
\item 
{We design a hybrid architecture that {unifies end-to-end trajectory planning and reactive control. Joint post-optimization of both branches reconciles long-term planning with short-term reactive adjustments, leading to improved closed-loop planning performance.}}
\item{We demonstrate the effectiveness of TakeAD on the Bench2Drive benchmark, 
{achieving an {12.50}\% improvement in Driving Score over the previous state-of-the-art (SOTA) method.}
}
\end{enumerate}
\section{Related Works}
\subsection{IL-based End-to-end Driving}
Current E2E AD primarily follows the IL paradigm, showing considerable potential. UniAD~\cite{uniad} integrates full-stack driving tasks into a single unified network, showcasing the strong potential of the E2E approach. 
VAD~\cite{vad} models the driving scene using a fully vectorized representation, improving computational efficiency. 
VADv2~\cite{vadv2} further introduces probabilistic planning to cope with the uncertainty nature of planning. 
DriveTransformer~\cite{drivetransformer} proposes a transformer-based task-parallel architecture that simplifies system complexity. 
Despite these advances, IL-based policies suffer from the gap between open-loop training and closed-loop evaluation. 
To bridge this, we propose a post-optimization framework leveraging expert takeover data, which integrates DAgger-based imitation with DPO-based preference optimization.

\subsection{Post-Optimization for End-to-end Driving}
{Recent work has explored enhancing E2E AD through post-optimization techniques to overcome the limitations of purely IL-based training~\cite{piwm,trajgen}.
RAD~\cite{rad} establishes a 3D Gaussian splatting-based reinforcement learning (RL) post-training paradigm. 
AutoVLA~\cite{autovla} combines vision-language-action modeling with adaptive reasoning, using supervised fine-tuning for fast/slow thinking followed by reinforcement fine-tuning for task-aligned planning.
Drive-R1~\cite{drive-r1} explores bridging scenario reasoning and motion planning by combining supervised chain-of-thought finetuning with RL.
Despite these advances, post-optimization in E2E AD still faces two fundamental challenges: what data should be used and how to model the optimization target (i.e., reward function). 
We address both by introducing a takeover data collection pipeline and employing DPO~\cite{dpo} for implicit reward modeling to learn expert preferences in disengagement scenarios.
\section{Methods}

\begin{figure*}[htbp]
    \centering
    \includegraphics[width=\textwidth]{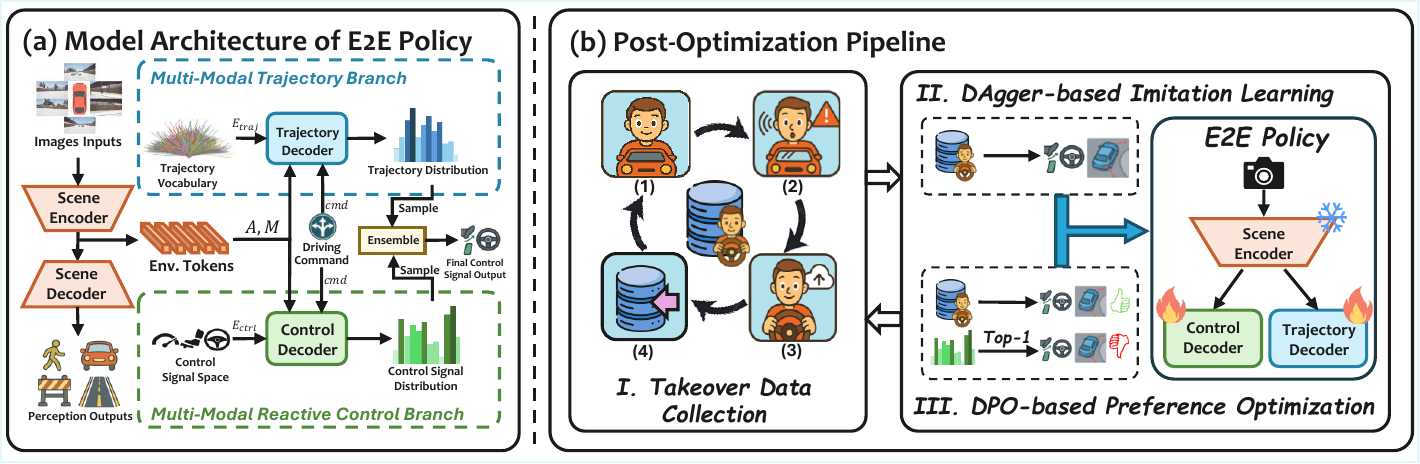}
    \vspace{-8mm}
    \caption{{\textbf{Overall framework of TakeAD.} 
    (a) Model architecture of TakeAD. 
    Multi-view images are encoded into environmental tokens and decoded into perception outputs, with parallel branches generating both long-term planning trajectories and instant control signals.
    (b) Post-optimization pipeline of TakeAD. 
    In the post-optimization phase, we adopt a multi-round iterative paradigm.  
    In each iteration, we sequentially perform expert takeover data collection, DAgger-style imitation, and preference optimization to expand the policy’s capability boundary based on high-quality takeover data. 
    Specifically, the expert takeover data collection process consists of: 
    (1) running the expert policy in “shadow mode” alongside the e2e policy; 
    (2) triggering a takeover when the e2e policy fails; 
    (3) letting the expert take control and record the intervention data;
    (4) storing expert takeover data.}}
    \label{fig2}
    \vspace{-6mm}
\end{figure*}
\subsection{End-to-End Driving Policy}
Our end-to-end driving policy consists of three components: Scene Encoder, Multi-modal Trajectory Branch, and Multi-modal Reactive Control Branch. Given multi-view images, {two parallel branches} produce probability distributions over predefined trajectory and control signal spaces.
A trajectory and corresponding control signal are sampled from their respective output distributions and integrated through an ensemble mechanism to generate the final vehicle control commands.

\textbf{\textit{Scene Encoder.}} 
The scene encoder adopts the VAD-style vectorized perception~\cite{vad}, encoding multi-view image sequences into map tokens $M\in R^{N_m\times C}$ and agent tokens $A\in R^{N_a\times N_{mod}\times C}$ via a series of attention operations. Here, $N_m$, $N_a$, $N_{mod}$ are the numbers of map elements, agents, and trajectory modes. {$C$ is the feature dimension.}
These tokens can be decoded into traffic element positions and agent trajectories.

\textbf{\textit{Multi-modal Trajectory Branch.}}
To model the uncertainty inherent in planning, we adopt a probabilistic planning~\cite{vadv2} approach. 
Specifically, we apply k-means clustering on the future ego trajectories from demonstrations to obtain $k$ cluster centers, thereby discretizing the trajectory space into a vocabulary $V_k$ of size $k$.

{We initialize learnable {trajectory} tokens $E_{traj}\in R^C$, which is augmented with navigation command features and positional encodings:
\begin{equation}
\label{equa4}
E_{traj}=E_{traj}+\text{MLP}(cmd)+\text{MLP}(E_v),
\end{equation}
where MLP denotes a multi-layer perceptron, $cmd$ is a one-hot encoding of the navigation instruction {$cmd$ is a one-hot encoding of the navigation instruction (in the benchmark we use, it includes six categories: Straight, Left, Right, LaneFollow, ChangeLaneLeft, ChangeLaneRight, and Void)}, and $E_v$ represents sine-based positional encodings computed from the spatial coordinates of each trajectory in $V_k$. 
The updated {trajectory} tokens $E_{traj}\in R^{k\times C}$ represent $k$ planning candidates enriched with high-level navigation intent and trajectory-space semantics.}
To incorporate scene context, {trajectory} tokens are fused with both agent tokens $A$ and map tokens $M$ through cross-attention (CA), {producing $E_{traj}^{agt}$ and $E_{traj}^{map}$}:
\begin{equation}
\label{equa6}
\begin{gathered}
    E_{traj}^{agt}=\text{CA}(E_{traj},A,A),\\
    E_{traj}^{map}=\text{CA}(E_{traj}^{agt},M,M).
\end{gathered}
\end{equation}
Finally, the probability distribution over the trajectory vocabulary $d_{traj}$ is {obtained by applying an MLP followed by a Sigmoid activation to the concatenated $E_{traj}^{agt}$ and $E_{traj}^{map}$:}
\begin{equation}
\label{equa7}
d_{traj}=\text{Sigmoid}[\text{MLP}(\text{Cat}(E_{traj}^{agt},E_{traj}^{map}))].
\end{equation}

\textbf{\textit{Multi-modal Reactive Control Branch.}}
Most existing E2E AD policies predict future waypoints at fixed time intervals~\cite{uniad,w4d}. 
Heuristic post-processing is typically applied to derive target velocity and steering angle, which are subsequently translated into low-level control signals using PID (Proportion Integration Differentiation) or LQR (Linear Quadratic Regulator) controllers~\cite{b2d}. 
While {long-term trajectory planning} contributes to stable and smooth driving, it often leads to delayed reactions in closed-loop deployment when encountering sudden events~\cite{tcp}. 
In contrast, control signals directly affect ego-vehicle behavior, allowing a reactive response to interactive situations. 
To this end, we implement a multi-modal control branch for {generating reactive vehicle control signals}.

Similar to the trajectory branch, we first discretize the control signal space. 
Specifically, {we discretize the control outputs—throttle, brake, and steering—into three separate action spaces with 5, 2, and 9 discrete actions respectively, following the value settings defined in Bench2Drive~\cite{b2d}.} 
Unlike the trajectory branch, we design a set of learnable control tokens $E_{ctrl}\in R^{N_c\times C}$, where $N_c$ represents the sum of the number of three discrete control signals. 
The control token undergoes similar interactions with the trajectory branches to obtain a probability distribution over control signals {$d_{ctrl}$}:
\begin{equation}
\label{equa8}
\begin{gathered}
E_{ctrl}^{agt} = \text{CA}(E_{ctrl}, A, A), \\
E_{ctrl}^{map} = \text{CA}(E_{ctrl}^{agt}, M, M), \\
d_{ctrl} = 
\text{Softmax}\left[\text{MLP}\left(\text{Cat}(E_{ctrl}^{agt}, E_{ctrl}^{map})\right)\right],
\end{gathered}
\end{equation}
{where Softmax denotes the softmax activation function.}

\textbf{\textit{Ensembling Trajectory and Control Signals.}}
As discussed above, predicting only trajectories struggles to handle interactive scenarios effectively, while directly generating control signals often suffers from frame inconsistency~\cite{tcp}. 
To leverage the advantages of both, we ensemble the long-term trajectory planning and reactive control prediction.
Trajectory facilitates stable driving in general scenarios, while the control signal further enhances performance in interactive scenarios.

During inference, the output trajectory $\tau_{plan}$ and control signal $c_{ctrl}=(throttle_{ctrl}, brake_{ctrl}, steer_{ctrl})$ are sampled from their respective probability distributions {$d_{traj}, d_{ctrl}$}. 
By default, a top-1 sampling strategy is employed.
Next, a PID controller converts $\tau_{plan}$ into control signal $c_{traj}=(throttle_{traj}, brake_{traj}, steer_{traj})$. The final output control signals are obtained by averaging the throttle and steering from both branches, while the braking value is taken as the logical OR (max) of the two:
\begin{equation}
    \label{equa10}
    \begin{gathered}
        throttle_{final}=\text{mean}(throttle_{ctrl},throttle_{traj}),\\
steer_{final}=\text{mean}(steer_{ctrl}, steer_{traj}),\\
brake_{final}=\text{max}(brake_{ctrl},brake_{traj}).
    \end{gathered}
\end{equation}
This hybrid output strategy, combining trajectory and control signals, enables more robust closed-loop planning.

\subsection{Takeover Data Collection}
In real-world autonomous driving systems, a takeover request is {triggered} to the human driver when the system encounters “critical events” that exceed its handling capability or preset safety thresholds, thereby ensuring driving safety. Inspired by this process, we propose a systematic pipeline to collect high-quality takeover data, efficiently filtering disengagement events from large-scale driving scenarios. 

\begin{figure}[!t]
\centering
\includegraphics[width=0.43\textwidth]{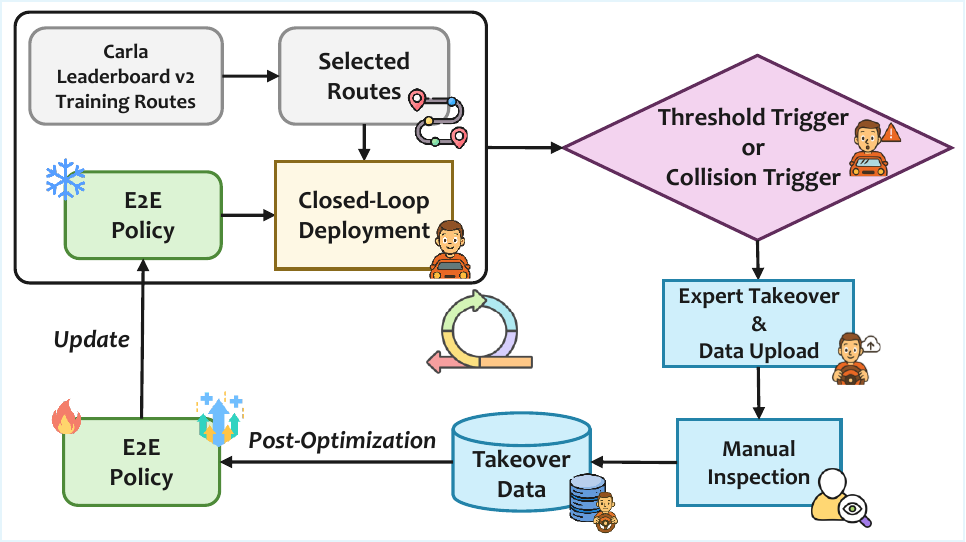}
\vspace{-2mm}
\caption{{Expert takeover data collection pipeline.}}
\label{fig3}
\vspace{-6mm}
\end{figure}
Specifically, we employ a rule-based expert policy, PDM-Lite~\cite{pdmlite}, to simulate the human expert. {PDM-Lite bypasses visual perception and directly accesses the ground-truth states (map topology, traffic participants kinematics, traffic signs, etc.) from the CARLA simulator, which allows PDM-Lite to achieve SOTA performance under perfect perception conditions on the CARLA Leaderboard 2.0.}

To accelerate algorithm iteration, we adopt a rule-based filtering strategy based on driving scores. Specifically, we select scenario types with an average driving score below 60, {where collisions are highly likely to occur.} Corresponding routes in the CARLA Leaderboard v2 training set are extracted, and our policy is executed in a closed-loop manner on these routes. Meanwhile, the expert policy runs concurrently in a “shadow mode.” The expert takeover is triggered when either of the following two conditions is met:

\begin{enumerate}
    \item Collision trigger: The expert policy detects a potential future collision risk for the ego vehicle;
    \item Threshold trigger: The absolute difference between the steering angle predicted by the model and that of the expert exceeds a preset threshold $\epsilon_{steer}=0.2$.
\end{enumerate}

The two conditions ensure correction of unsafe behaviors or major heading deviations.
Once the takeover condition is triggered, the expert takes control of the vehicle. 
A fixed takeover duration of 2 seconds is used to ensure the vehicle can recover to a stable driving state.
During the takeover period, sensor inputs, control outputs from both the model and the expert are recorded. 
{The manual inspection aims to discard the small portion of data where the expert exhibits abnormal behaviors (e.g., collisions, route deviations, or getting stuck). This process can also be automated by leveraging the simulator’s infraction logs (e.g., Collision, Off-road events) and speed records.}
Finally, we obtain the expert takeover dataset $D_{takeover}$ for subsequent post-optimization.

\subsection{Pre-training Based on Imitation Learning}
 The goal of the pre-training stage is to fit the distribution of expert demonstrations via imitation learning. Motivated by VAD~\cite{vad}, the loss function consists of five tasks: detection, mapping, prediction, long-term trajectory planning, and reactive control signal {generation}. {Following the same definitions as in \cite{vad}, the detection, mapping, and prediction losses are further divided into classification and regression losses represented as the following superscript $cls$ and $reg$ respectively}:
\begin{equation}
    \label{equa13}
    \begin{gathered}
        L_{det}=L_{det}^{cls}+L_{det}^{reg},\\
        L_{map}=L_{map}^{cls}+L_{map}^{reg},\\
        L_{pred}=L_{pred}^{cls}+L_{pred}^{reg},
    \end{gathered}
\end{equation}
{where the losses with the subscript $det$ and $map$ learn attributes of traffic elements, and the losses with the subscript $pred$ focus on predicting multi-modal trajectories of agents.}

The long-term trajectory planning loss $L_{traj}$ and the instantaneous control {generation} loss $L_{ctrl}$ are defined as the KL divergence {($D_{KL}$)} between the predicted distributions and ground truth distributions:
\begin{equation}
    \label{equa14}
    \begin{gathered}
        L_{traj}=D_{KL}(d_{traj}||d_{traj}^{expert}),\\
        L_{ctrl}=D_{KL}(d_{ctrl}||d_{ctrl}^{expert}),
    \end{gathered}
\end{equation}
where $d_{traj/ctrl}$ and $d_{traj/ctrl}^{expert}$ denote the predicted distributions and the expert distributions for trajectory and control, respectively. 
{The pre-training procedure is divided into 3 stages to ensure training stability.}
Each stage is designed to update a specific component of the model: the perception backbone, the trajectory branch, and the control branch, respectively. 

\vspace{-1.5mm}

\subsection{Post-optimization Algorithm Based on DAgger and DPO}
Given the pre-trained E2E model, we expect to mitigate the open-loop gap and improve the closed-loop performances by leveraging the expert takeover data. Given the inherent challenges in explicitly modeling reward functions from takeover data, we naturally formulate expert takeover actions as pairwise preferences. This motivates our adoption of DPO, which bypasses explicit reward modeling by directly optimizing policies using relative preference comparisons.
During post-optimization, 
{we update both the trajectory and control branches to maintain consistency between long-term behaviors and short-term adjustments.}
We perform DAgger-based imitation learning~\cite{DAgger} prior to DPO-based preference optimization, in order to initially enhance the policy’s ability to handle {disengagement} states.

\textbf{\textit{DAgger-based Imitation Learning.}}
As an online IL algorithm, DAgger iteratively collects data under the current policy and aggregates it with the original dataset to update the policy. By including states encountered during policy execution, especially those outside the original distribution, DAgger expands data coverage and enhances model performance through interaction with the environment.

{In our framework, by iteratively aggregating the takeover dataset $D_{takeover}$—which contains critical samples representing the capability boundary of the current policy—into the demonstration dataset $D_{demo}$, DAgger explicitly exposes the policy to recovery scenarios beyond its initial training distribution, thereby enhancing its robustness while maintaining the fundamental driving capability in common scenarios.}
We adopt an imitation learning loss $L_{\text{DAgger}}$, which minimizes the KL divergence between the predicted distribution {{$d_{ctrl, traj}$}} and the distribution of the aggregated dataset {$d_{DAgger}$}:
\begin{equation}
    \label{equa17}
    L_{\text{DAgger}}=D_{KL}(d_{ctrl, traj}||d_{DAgger}).
\end{equation}

\textbf{\textit{DPO-based Preference Optimization.}} 
To further align the behavior of the policy with expert demonstrations in disengagement scenarios, we construct preference data based on the takeover dataset $D_{takeover}$. Unlike typical preference learning approaches in LLMs~\cite{dpo}, where the preference dataset is constructed in advance, we dynamically generate preference pairs during training based on the latest model predictions.

{Here, we use SimPO~\cite{simpo}, a variant version of DPO that better aligns model generation without requiring a reference model.
The original objective function of SimPO is given by:
\begin{align}
\label{equa12}
L_{\mathrm{SimPO}}(\pi_\theta,D) = {} &
- \mathbb{E}_{(x, y_w, y_l) \sim D} \Bigg[
    \log \sigma\Big(
        \frac{\beta}{|y_w|}\log\pi_\theta(y_w|x) \notag \\
    & \quad
        -\frac{\beta}{|y_l|}\log\pi_\theta(y_l|x)-\gamma
    \Big)
\Bigg],
\end{align}
where {$x$ is the input to the policy $\pi_\theta$, $y_w$ and $y_l$ denote the preferred and dispreferred outputs, $D$ is the preference dataset, $\sigma$ is the sigmoid activation function,} $\gamma$ is a hyperparameter introduced to amplify the difference between $y_w$ and $y_l$, $\beta$ is a constant that controls the scaling of the reward difference. 
For each sample in the takeover dataset $D_{takeover}$, the expert takeover action is treated as the preferred action, while the policy's output is regarded as the dispreferred one.} 
To prevent unnecessary updates when the expert action has already become the most probable prediction,
 we revise the {preference optimization (PO)} loss as:
\begin{equation}
    \label{equa18}
    L_{\mathrm{PO}}(\pi_\theta)=L_{\mathrm{SimPO}}(\pi_\theta,D_{takeover})+\log\sigma\left(-\gamma\right).
\end{equation}
{Here, the sequence length $|y_w|=|y_l|=1$.} Since the dispreferred action $y_l$ is the one with the highest predicted probability from the model, i.e., $y_l=\arg\max_y\pi_\theta(y|x),$ we have $\beta log\pi_\theta(y_w|x)-\beta log\pi_\theta(y_l|x)<0.$ A compensation term $\log\sigma(-\gamma)$ is added to the loss function, ensuring that the minimum value $L_{PO}$ is zero.

\textbf{\textit{Multi-round Fine-tuning with DAgger and SimPO.}}
As shown in Algorithm \ref{alg:dagger_dpo}, we alternate between DAgger and SimPO to iteratively update the policy during the post-optimization stage. In each iteration, we use the latest policy to collect a new takeover dataset and apply DAgger and SimPO in sequence for policy refinement. The purpose of DAgger is to expand the capability boundary of the policy by enabling it to imitate expert actions in {disengagement} states, thereby equipping the policy with an initial ability to generate preferred actions. This provides a solid foundation for subsequent preference optimization. The goal of SimPO is to model implicit rewards from the preference data, thereby aligning the policy’s behavior more closely with expert preferences in disengagement scenarios.

 \section{Experiments}
 
\subsection{Experiment Setup}
\textbf{\textit{Dataset and Benchmark.}}
We adopt the Bench2Drive closed-loop benchmark (B2D)~\cite{b2d} to train and evaluate TakeAD. 
During pre-training, the model is trained on the B2D-base dataset, which is collected by a privileged model-based RL agent, Think2Drive~\cite{think2drive}. 
We follow the official B2D closed-loop evaluation protocol, which consists of 220 challenging routes covering 44 types of interactive scenarios, under diverse weather conditions and across multiple towns.

\begin{algorithm}[ht]
\caption{Multi-round Post-optimization Algorithm}
\label{alg:dagger_dpo}

\begin{algorithmic}[1]
\STATE \textbf{Input:} Pretrained policy $\pi_{pre}$, expert demonstrations $D_{demo}$
\STATE \textbf{Output:} Preference-aligned policy $\pi_{PO}$

\STATE $\pi_{current} \leftarrow \pi_{pre}$
\FOR{$i = 1, 2, \dots, I$}
    \STATE Collect takeover dataset $D_{takeover}^i$ using $\pi_{current}$
    \STATE Merge $D_{takeover}^i$ with $D_{demo}$ to form $D_{DAgger}^i$
    \FOR{$j = 1, 2, \dots, N$}
        \STATE Sample data from $D_{DAgger}^i$
        \STATE Compute $L_{\text{DAgger}}$ according to (\ref{equa17})
        \STATE Backpropagate gradients and update $\pi_{current}$
    \ENDFOR

    \FOR{$k = 1, 2, \dots, N$}
        \STATE Sample data from $D_{takeover}^i$
        \STATE Obtain policy output distribution $\pi_{current}(y \mid x)$
        \STATE Calculate $y_l = \max_i \pi_{current}(y_i \mid x)$
        \STATE Compute $L_{\text{PO}}$ according to (\ref{equa18})
        \STATE Backpropagate gradients and update $\pi_{current}$
    \ENDFOR
\ENDFOR
\STATE \textbf{return} $\pi_{PO} \leftarrow \pi_{current}$
\end{algorithmic}
\end{algorithm}

\textbf{\textit{Evaluation Metrics.}}
We use the following metrics to assess driving performance. 
\textbf{Driving Score (DS)} combines Route Completion and Infraction Score, reflecting both task success and rule compliance. 
\textbf{Success Rate (SR)} indicates the proportion of routes completed on time without violations. 
\textbf{Efficiency} measures the ego vehicle's speed relative to surrounding traffic, encouraging progressiveness without aggression. 
\textbf{Comfortness} evaluates the smoothness of the driving trajectory.

\textbf{\textit{Implementation Details.}}
We adopt a VAD-style~\cite{vad} perception module using a ResNet-50 backbone to extract image features from six surround-view cameras with a resolution of 768×1280, covering a perception range of 60m × 30m. The trajectory vocabulary size is set to $k=4096$. During pre-training, we use two NVIDIA A800 GPUs with a total batch size of 8. The perception, trajectory and control pre-training stages {consist of} 8, 3, and 3 epochs respectively, using a cosine annealing learning rate schedule with an initial learning rate of 2e-4. During post-optimization, we collected {8596, 6106, 5799, 6395 and 6078} samples in five successive iterations using the latest policy. 
In each iteration, {we merge all current and previous takeover data with the original demonstration dataset $D_{demo}$}, which is then used for DAgger-based IL. Preference optimization is conducted only on the takeover data collected in the current iteration.
{\textbf{All training data for post-optimization are collected on the training routes of CARLA Leaderboard v2, which are distinct from the testing routes.}}
The post-optimization stage is performed using six NVIDIA A800 GPUs with a total batch size of 48. DAgger and PO are trained for {1} and 10 epochs, respectively, with initial learning rates of 5e-5 and 1e-6. {Each post-optimization round takes about 18 hours (15h for DAgger, 3h for DPO).} In (\ref{equa18}), we set $\beta=0.1, \gamma=0.1$. 

Additionally, during closed-loop evaluation, we observe an inertia issue common in imitation learning methods, where the ego vehicle tends to remain stationary after a stop~\cite{il-limit}. Inspired by prior work~\cite{transfuser}, we adopt a safety creeping strategy: when no leading vehicle is detected and the ego vehicle remains still for 2.5 seconds, a throttle of 0.7 is applied for 1 second to resume motion. 
\subsection{Results and Analysis}
\begin{table*}[htbp]
\centering

\small
\caption{\textbf{{Closed-loop and Open-loop Results of E2E-AD Methods in Bench2Drive}}. 
\vspace{-3mm} 
\label{tab: b2d}}
\resizebox{0.74\linewidth}{!}
{
\begin{threeparttable}
\begin{tabular}{lc|>{\columncolor[gray]{0.9}}c>{\columncolor[gray]{0.9}}ccc|c}
\toprule
\multirow{2}{*}{\textbf{Methods}} & \multirow{2}{*}{\textbf{Reference}} & \multicolumn{4}{c|}{\textbf{Closed-loop Metric}} & \multicolumn{1}{c}{\color{gray} \textbf{Open-loop Metric}} \\ 
\cmidrule{3-7} 
& & DS $\uparrow$  & SR (\%) $\uparrow$ & Efficiency $\uparrow$ & Comfortness $\uparrow$ & \color{gray} Avg. L2 (m) $\downarrow$\\ 
\midrule

TCP*~\cite{tcp} & NeurIPS 22 & 40.70 & 15.00 & 54.26 & \underline{47.80} & \color{gray} 1.70 \\ 
ThinkTwice*~\cite{thinktwice} & CVPR 23 & 62.44 & 31.23 & 69.33 & 16.22 & \color{gray} 0.95\\
DriveAdapter*~\cite{DriveAdapter} & ICCV 23 & \underline{64.22} & 33.08 & 70.22 & 16.01 & \color{gray} 1.01 \\ 
\midrule
\midrule
                                 
UniAD~\cite{uniad} & CVPR 23 & 45.81 & 16.36 & 129.21 & 43.58 & \color{gray} \underline{0.73} \\
VAD~\cite{vad} & ICCV 23 & 42.35 & 15.00 & 157.94 & 46.01 & \color{gray} 0.91 \\
MomAD~\cite{momad} & CVPR 25 & 47.91 & 18.11 & 174.91 & \textbf{51.20} & \color{gray} 0.82 \\
DriveTransformer-Large~\cite{drivetransformer} & ICLR 25 & 63.46 & \underline{35.01} & 100.64 & 20.78 & \color{gray} \color{gray} \textbf{0.62} \\ 

\midrule
TakeAD w.o. post (ours) & - & 61.82 & 25.35 & \textbf{194.93} & 20.40 & \color{gray} 0.87 \\ 
TakeAD (ours) & - & \textbf{71.39} & \textbf{40.83} & \underline{193.30} & 22.89 & \color{gray} 0.91 \\ 
\midrule
\color{gray} \textit{PDM-Lite}$^\dag$~\cite{pdmlite} & - & \color{gray} \textit{96.25} & \color{gray} \textit{83.49} & \color{gray}\textit{22.33}  & \color{gray}\textit{213.47}  & \color{gray}{-} \\
\bottomrule

\end{tabular}
\end{threeparttable}
}
\vspace{-2mm}
\end{table*}

\begin{table*}[htbp]
\centering
\caption{{\textbf{{Multi-Ability Results of E2E-AD Methods in Bench2Drive.}}}
\vspace{-3mm}
\label{tab: ability}}
\resizebox{0.81\linewidth}{!}{
\begin{threeparttable}
\begin{tabular}
{lc|ccccc|>{\columncolor[gray]{0.9}}c}
\toprule
\multirow{2}{*}{\textbf{Methods}} & \multirow{2}{*}{\textbf{Reference}} & \multicolumn{5}{c}{\textbf{Ability} (\%) $\uparrow$}  \\ 
\cmidrule{3-8} 
& & \multicolumn{1}{c}{Merging} & \multicolumn{1}{c}{Overtaking} & \multicolumn{1}{c}{Emergency Brake} & \multicolumn{1}{c}{Give Way} & Traffic Sign & \textbf{Mean} \\ 
\midrule

TCP*~\cite{tcp} & NeurIPS 22 & 17.50  & 13.63 & 20.00 & 10.00 & 6.81 & 13.59 \\
ThinkTwice*~\cite{thinktwice} & CVPR 23 & 13.72 & 22.93 & 52.99 & \textbf{50.00} & 47.78 & 37.48 \\ 
DriveAdapter*~\cite{DriveAdapter} & ICCV 23 & 14.55 & 22.61 & \underline{54.04} & \textbf{50.00} & \underline{50.45} & 38.33 \\
\midrule
\midrule

UniAD~\cite{uniad} & CVPR 23 & 12.16 & 20.00 & 23.64 & 10.00 & 13.89 & 15.94 \\ 
VAD~\cite{vad} & ICCV 23 & 7.14 & 20.00 & 16.36 & 20.00 & 20.22 & 16.75 \\ 
DriveTransformer-Large~\cite{drivetransformer} & ICLR 25 & 17.57 & \underline{35.00} & 48.36 & \underline{40.00} & \textbf{52.10} & \underline{38.60} \\ 

\midrule

TakeAD w.o. post (ours) & - & \underline{24.68} & 13.33 & 36.67 & \underline{40.00} & 31.91  & 29.32 \\ 
TakeAD (ours) & - & \textbf{30.77} & \textbf{35.56} & \textbf{56.67} & \textbf{50.00} & 42.02  &  \textbf{43.00} \\ 
\midrule
\color{gray} \textit{PDM-Lite}$^\dag$~\cite{pdmlite} & - & \color{gray} \textit{82.05} & \color{gray} \textit{91.11} & \color{gray}\textit{81.67}  & \color{gray}\textit{80.00}  & \color{gray}\textit{70.97} & \color{gray}\textit{81.16} \\
\bottomrule
\end{tabular}
\begin{tablenotes}
     \item[*] indicates expert feature distillation methods with the target waypoint as navigation. Other methods (in normal text) have no access to the expert feature and rely on one-hot commands for navigation.

     \item[$^\dag$] Reproduced results using the open-sourced code of PDM-Lite, the privileged expert in our post-optimization process.
\end{tablenotes}
\end{threeparttable}
}
\vspace{-6mm}
\end{table*}

\textit{1) \textbf{Quantitative Comparison.}} {We conduct comprehensive experiments in closed-loop scenarios to evaluate TakeAD’s driving performance.}

{\textbf{SOTA Performance in Challenging Closed-Loop Evaluations.}} Previous studies~\cite{admlp,b2d} have shown that open-loop metrics primarily serve as indicators of model convergence, whereas closed-loop metrics offer more reliable evaluations of driving capabilities. Therefore, we focus on the closed-loop metric. As shown in Table \ref{tab: b2d}, TakeAD achieves SOTA performance in terms of both DS and SR on the Bench2Drive closed-loop benchmark. Specifically, {after post-optimization, }TakeAD surpasses the previous SOTA DriveAdapter~\cite{DriveAdapter}, which relies on expert feature distillation and precise target waypoints for navigation. TakeAD improves the DS by {7.17} and the SR by {7.75}\% over DriveAdapter. Compared with the backbone method VAD~\cite{vad}, TakeAD achieves a {68.57}\% gain in DS and a {25.83}\% gain in SR. Additionally, TakeAD reaches the highest Efficiency score, indicating an effective and proactive driving strategy. 
{TakeAD's lower comfort metric stems from its prioritization of safety, as it executes necessary braking and evasive maneuvers to ensure route completion in interactive scenarios. Under the current benchmark setting, safe driving behaviors often come at the cost of comfort; thus, we believe that comfort comparisons are meaningful only among models with comparable driving performance.}
Table \ref{tab: ability} presents multi-ability evaluations. 
TakeAD achieves an average capability score of {43.00}\%, outperforming previous methods.

\begin{table}[h]
\renewcommand{\arraystretch}{1.0}
\vspace{-4mm}
\caption{
{Zero-shot Closed-loop Results in DOS.
DOS\_01 = Parked Cars, DOS\_02 = Sudden Brake, DOS\_03 = Left Turn, DOS\_04 = Red Light Infraction.
Only the driving score is reported.}
}
\vspace{-4mm}
\label{dos}
\setlength\tabcolsep{2pt}
\begin{center}
\resizebox{0.80\linewidth}{!}{\begin{tabular}{c|c|c|c|c|c}
\toprule
\multirow{2}{*}{\textbf{Methods}} & \multicolumn{5}{c}{\textbf{Driving Score (DS) in DOS $\uparrow$}} \\
\cmidrule{2-6}
& DOS\_01 & DOS\_02 & DOS\_03 & DOS\_04 & Average \\
\cmidrule{1-6}
UniAD \cite{uniad} & \underline{66.00} & 69.32 & 73.18 & \underline{75.87} & \underline{71.09} \\
VAD \cite{vad} & 58.46 & 57.84 & 61.16 & 65.88 & 60.84 \\
MomAD \cite{momad} & 53.90 & 63.36 & 56.00 & 55.06 & 57.08 \\
LMDrive \cite{lmdrive}& 64.00 & 62.00 & \underline{73.65} & 58.54 & 64.53 \\
TakeAD w.o. post(ours) & 65.72 & \underline{75.00} & 64.32 & 67.65 & 68.17 \\
TakeAD(ours) & \textbf{71.54} & \textbf{79.01} & \textbf{76.41} & \textbf{79.71} & \textbf{76.67} \\
\hline 
\end{tabular}}
\end{center}
\vspace{-4mm}
\end{table}

{\textbf{Robust Zero-Shot Generalization in Complex Interactive Scenarios.}
To further assess the generalization ability of TakeAD under out-of-distribution (OOD) conditions, we conduct additional experiments on the Driving in Occlusion Simulation (DOS) benchmark \cite{reasonnet}.
As reported in Table \ref{dos}, TakeAD exhibits strong performance even before post-optimization and achieves the highest driving score after post-optimization, verifying that this stage effectively enhances the model’s generalization ability.}

\textit{2) \textbf{Ablation Study.}} We conduct ablation experiments to validate the effectiveness of each component.

\begin{table}[htbp]
\vspace{-2mm}
\centering
\caption{
\textbf{Ablation Study on control branch and safety creeping.} MMCB and SC denote Multi-Modal Control Branch and Safety Creeping. IS, RC and TO refer to Infraction Score, Route Completion and Time Out.
\vspace{-2mm}
\label{tab: ablation1}}
\resizebox{0.90\linewidth}{!}{
\begin{tabular}{c|cc|ccccc} 
\toprule
\multirow{2}{*}{\textbf{ID}} & \multicolumn{2}{c|}{\textbf{Settings}} & \multicolumn{5}{c}{\textbf{Closed-loop Metric}} \\ 
\cmidrule{2-8}
& MMCB & SC & DS $\uparrow$ & SR (\%) $\uparrow$ & IS $\uparrow$ & RC (\%) $\uparrow$ & TO (\%) $\downarrow$\\ 
\cmidrule{1-8}
1 & $\times$ & $\times$ & 53.20 & 24.20 & 0.73 & 75.96 & 31.96 \\
2 & \checkmark & $\times$ & 56.28 & \textbf{25.57} & \textbf{0.77} & 76.65 & 28.76 \\
3 & \checkmark & \checkmark & \textbf{61.82} & 25.35 & 0.67 & \textbf{91.31} & \textbf{14.29} \\
\bottomrule
\end{tabular}}
\end{table}

\textbf{\textit{Effectiveness of the Multi-modal Control Branch.}} 
As shown in Table \ref{tab: ablation1}, comparing ID1 and ID2, the multi-modal control branch improves DS by 5.79\% and IS by 5.48\%, along with slight improvements in SR and RC. This demonstrates that predicting control signals enhances the policy’s performance in interactive scenarios.


\textbf{\textit{Effectiveness of the Safety Creeping Mechanism.}}
In Table \ref{tab: ablation1}, the TO metric refers to the percentage of test routes (out of 220) that fail due to exceeding the time limit. 
Comparing ID2 and ID3, the safety creeping mechanism notably reduces the time-out rate, improving RC and DS, and mitigating the inertia issue observed in imitation-based policies.


\begin{table}[htbp]
\vspace{-3mm}
\centering
\caption{
\textbf{{Ablation Study on Different Post-Optimization Strategies. All experiments use the hybrid output strategy.}}
\vspace{-2mm}
\label{tab:ablation2}}
\resizebox{0.82\linewidth}{!}{
\begin{tabular}{c| cc| cc| cc}
\toprule
\multirow{2}{*}{\textbf{ID}} & 
\multicolumn{4}{c|}{\textbf{Post-optimization strategies}} &
\multicolumn{2}{c}{\textbf{Closed-loop Metric}} \\
\cmidrule(lr){2-7}
& \multicolumn{2}{c|}{Trajectory Branch} & \multicolumn{2}{c|}{Control Branch} & \multirow{2}{*}{DS $\uparrow$} & \multirow{2}{*}{SR (\%) $\uparrow$} \\
\cmidrule(lr){2-3}\cmidrule(lr){4-5}
& DAgger & DPO & DAgger & DPO &  &  \\
\midrule
1 & $\times$ & $\times$ & $\times$ & $\times$ & 61.82 & 25.35 \\
2 & \checkmark & $\times$ & \checkmark & $\times$ & 63.64 & 29.82 \\
3 & $\times$ & \checkmark & $\times$ & \checkmark & 63.44 & 33.18 \\
4 & \checkmark & \checkmark & $\times$ & $\times$ & 64.36 & 33.18 \\
5 & $\times$ & $\times$ & \checkmark & \checkmark & \underline{65.19} & \underline{34.10} \\
6 & \checkmark & \checkmark & \checkmark & \checkmark & \textbf{65.85} & \textbf{34.88} \\
\bottomrule
\end{tabular}}
\end{table}

\textbf{\textit{Comparison of Post-optimization Strategies.}}
Table \ref{tab:ablation2} compares the impact of different post-optimization strategies on policy performance. 
{Applying post-optimization to either the trajectory or control branch (ID4 and ID5) notably improves driving performance. Fine-tuning the control branch yields higher gains (DS ↑3.99, SR ↑8.75\%) than the trajectory branch (DS ↑3.37, SR ↑7.83\%). This improvement is attributed to the compact nature and superior inter-action distinguishability of the control signal space, which facilitates preference learning under limited data, and to the short-term characteristics of takeover maneuvers that are more effectively captured in the reactive control. Jointly optimizing both branches (ID6) yields further gains, as it enforces consistency between long-horizon planning and short-term reactive control, reducing behavioral conflicts. 
Comparing ID1–ID3, single-stage optimization proves insufficient. DAgger alone provides limited improvement, as IL increases preferred actions but fails to suppress undesired ones. Similarly, DPO alone yields modest gains due to the pre-trained model’s insufficient capability in handling disengagement scenarios. Our two-stage approach (ID6) achieves the best results (DS ↑6.52\%, SR ↑9.53\%), effectively using DAgger as a warm-up to facilitate subsequent preference optimization.}

\textbf{\textit{Multi-round Iterative Post-optimization Results.}}
Fig. \ref{fig4} illustrates the impact of multiple iterations of our post-optimization strategy on the E2E AD policy. Up to the fourth iteration, the policy continues to improve, achieving {15.48}\% and {15.48}\% gains in DS and SR, respectively. These improvements demonstrate the significant potential of the proposed post-optimization framework. However, after the fourth iteration, the performance gain saturates due to the limitations of the VAD base model. 
Specifically, our base model lacks inputs such as navigation waypoints or traffic light recognition, placing it at an inherent disadvantage in complex scenarios like intersections and junctions. 
Further improving the model architecture remains an important direction for future work.

\begin{figure}[htp]
\vspace{-4mm}
\centering
\includegraphics[width=0.4\textwidth]{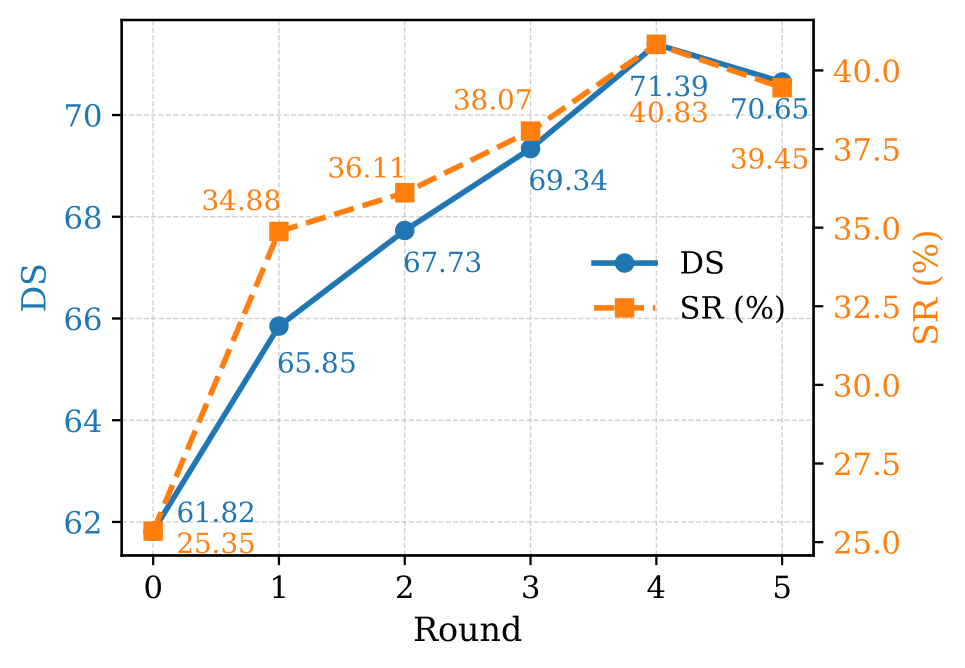}
\vspace{-4mm}
\caption{{Multi-round Iterative Post-optimization Results.}}
\label{fig4}
\end{figure}
\vspace{-2mm}

\textit{3) \textbf{Qualitative Visualization.}}  
Fig. \ref{fig5} compares the closed-loop performance of the policy before and after four iterations of post-optimization. The fine-tuned policy shows clear improvements over the pre-trained policy in challenging scenarios like intersection turns and emergency avoidance, successfully preventing potential collisions.

\section{Conclusion}
In this work, we propose TakeAD, a novel post-optimization framework for E2E AD that leverages expert takeover data from {disengagement} scenarios. Our framework combines DAgger-based IL and DPO-based preference optimization. DAgger provides an initial ability to handle disengagement cases, while DPO further aligns the policy with expert preferences. Experiments on the Bench2Drive closed-loop benchmark demonstrate the effectiveness of our approach, achieving SOTA performance. 

{However, the inertia issue persists, particularly at signalized intersections, due to a limitation of the base model: it detects traffic lights as objects but does not explicitly recognize their states, making it difficult to associate signals with actions. Similar to Transfuser~\cite{transfuser} and MoMAD~\cite{momad}, TakeAD still relies on a rule-based safety creeping strategy. In future work, we plan to address this issue through post-optimization on more advanced base models and investigate the continuous learning techniques for end-to-end autonomous driving.}

\begin{figure*}[htb]
    \centering
    \includegraphics[width=\textwidth]{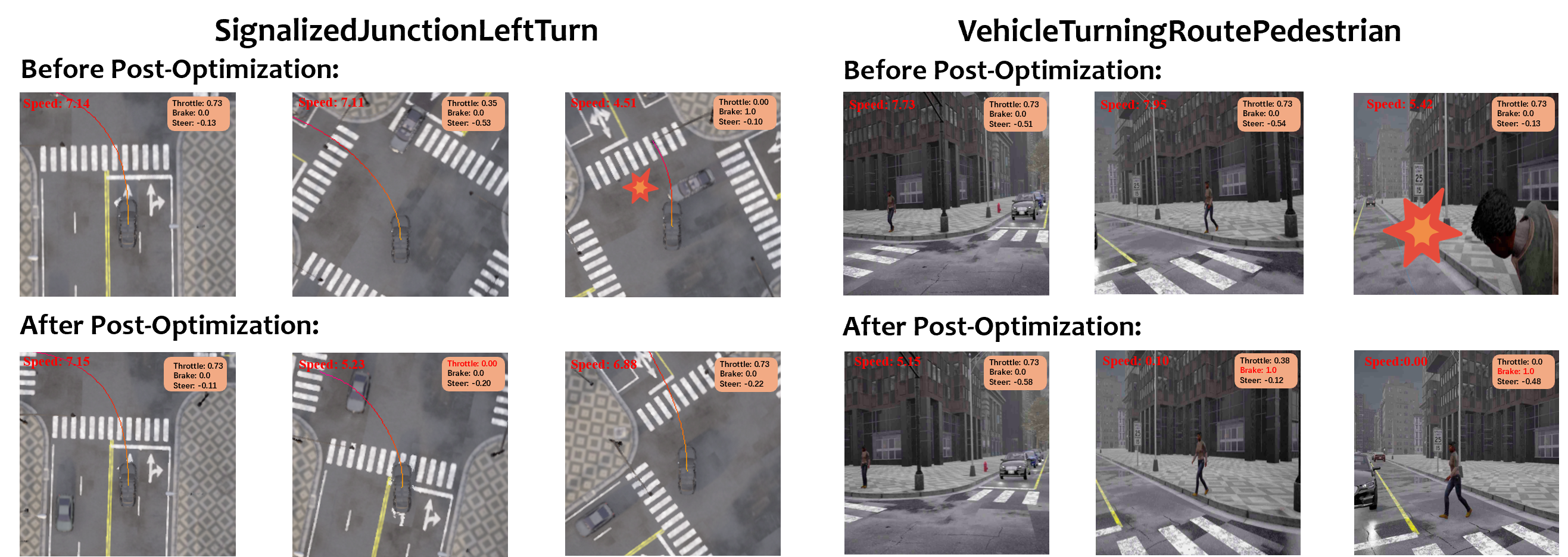}
    \caption{{Visualization of our E2E AD policy before and after post-optimization in typical scenarios. {Note: In CARLA, both throttle and brake can be greater than zero at the same time, but only the brake will take effect in this case.}}}
    \label{fig5}
    \vspace{-6mm}
\end{figure*}

\vfill

\end{document}